\documentclass{article}

    \PassOptionsToPackage{numbers, compress}{natbib}
\usepackage{paralist}

\usepackage[preprint]{neurips_2024}

\usepackage[utf8]{inputenc} % allow utf-8 input
\usepackage[T1]{fontenc}    % use 8-bit T1 fonts
\usepackage{hyperref}       % hyperlinks
\usepackage{url}            % simple URL typesetting
\usepackage{booktabs}       % professional-quality tables
\usepackage{amsfonts}       % blackboard math symbols
\usepackage{nicefrac}       % compact symbols for 1/2, etc.
\usepackage{microtype}      % microtypography
\usepackage{xcolor}         % colors
\usepackage{graphicx}
\usepackage{caption, subcaption}  %%{subfigure}
\usepackage{multirow}
\usepackage{algorithm}
\usepackage[noend]{algorithmic}
\usepackage{amsmath}
\usepackage{array} % required for text wrapping in tables
\usepackage[most]{tcolorbox}
\usepackage{pifont, arydshln}
\usepackage{hyperref}

\hypersetup{
    colorlinks,
  citecolor=magenta,
  linkcolor=red,
  urlcolor=blue           % color of external links
}
\usepackage{numprint}

\usepackage{amssymb}
\usepackage{mathtools}
\usepackage{amsthm}
\usepackage{enumitem}
\usepackage[capitalize,noabbrev]{cleveref}
\usepackage{wrapfig}
\usepackage[normalem]{ulem} % for strikeout font
\usepackage{blindtext}%
\usepackage[textsize=tiny]{todonotes}

\presetkeys{todonotes}{fancyline}{}

\newcommand{\dset}{\textsc{SugarCrepe++}\xspace}
\definecolor{airforceblue}{rgb}{0.36, 0.54, 0.66}

\title{Sensitivity of Generative VLMs to Semantically and Lexically Altered Prompts}
\author{%
  Sri Harsha Dumpala \quad
  Aman Jaiswal \quad
  Chandramouli Sastry \\
  {\bf Evangelos Milios} \quad
  {\bf Sageev Oore} \quad
  {\bf Hassan Sajjad} \\
  Dalhousie University, Canada.
}

\begin{document}

\maketitle

\begin{abstract}
Despite the significant influx of prompt-tuning techniques for generative vision-language models (VLMs), it remains unclear how sensitive these models are to lexical and semantic alterations in prompts. In this paper, we evaluate the ability of generative VLMs to understand lexical and semantic changes in text using the \dset dataset. We analyze the sensitivity of VLMs to lexical alterations in prompts without corresponding semantic changes. Our findings demonstrate that generative VLMs are highly sensitive to such alterations. Additionally, we show that this vulnerability affects the performance of techniques aimed at achieving consistency in their outputs.
\end{abstract}
%------------------------------------------------------------
\section{Introduction}
\label{sec:intro}
Vision-language models (VLMs) have achieved impressive performance across a wide range of vision and language downstream tasks~\cite{radford2021learning, alayrac2022flamingo, LiangWDLZ0ZVM23, imagen}. Despite this success, recent works have shown that VLMs lack compositional understanding and often struggle with reasoning about even simple spatial relationships or attribute attachments \cite{thrush2022winoground, yuksekgonul2023and, kamath2023text, hsieh2023sugarcrepe, kamath2023s}. Additionally, further research indicates that VLMs have difficulty comprehending simple lexical and semantic alterations~\cite{dumpala2024sugarcrepe++}. Most of these previous studies have focused on analyzing CLIP-based models \cite{radford2021learning, singh2022flava, schuhmann2022laion} to uncover the vulnerabilities of VLMs. In this work, we analyze the ability of generative VLMs, such as BLIP \cite{li2022blip}, BakLLavA \cite{liu2024improved}, and GPT-4o, to understand lexical and semantic alterations in input text/prompts using the recently released \dset dataset~\cite{dumpala2024sugarcrepe++}. %BLIP2~\cite{li2023blip2}

Prompt-tuning of generative VLMs has garnered significant research interest in recent times \cite{shu2022test, khattak2023maple, khattak2023self, abdul2024align, gu2023systematic, RoyE24_iclr}. Most of these works focus on carefully selecting a prompt template, specific to a downstream task, that works best for a given VLM. 
However, it is crucial to understand how lexical and semantic alterations to these prompts affect the output of VLMs. While some recent works have evaluated the impact of adversarial examples on VLM performance \cite{ZhangYS22_acm, ZhaoPDYLCL23_neurips, LuoGL024}, there are no works that analyze the prompt sensitivity of generative VLMs.
In this work, we analyze the sensitivity of VLMs to lexical variations (noticeable by humans) in input prompts that do not alter the semantic meaning. % Recent studies also analyzed consistency in the output of VLMs but these studies were confined to the downstream task of image classification.

Recent works have also evaluated the consistency in the output of large language models (LLMs) \cite{jiang2023llm, 0002WSLCNCZ23_iclr, ChenPPPZBC24_tmlr, sosareasoning, sivarajkumar2024empirical}, which is critical for deploying LLMs in real-time applications. One of the main focus of these works is to evaluate the self-consistency of LLMs~\cite{wang2023towards, tamkintask} i.e., to evaluate the consistency of LLMs by sampling multiple explanations and answers from the model. In this work, we analyze the consistency of VLMs using paraphrases of prompts. To the best of our knowledge, no prior work has assessed the consistency of generative VLM outputs. In particular, we analyze the output consistency of generative VLMs under two different settings: 1) inter-model (consistency across an ensemble of different VLMs) and 2) intra-model (consistency across prompts for the same model) consistency.

 The main contributions of this work are as follows:
 \begin{enumerate}
 \item We evaluate the sensitivity of generative VLMs to various lexical variations in the input prompt. Additionally, the main task is based on \dset, which requires VLMs to have a strong understanding of lexical and semantic alterations in text in order to achieve better performance.
 \item Evaluate the consistency in the output of VLMs. Here we evaluate two different approaches, namely, ensemble of models, and ensemble of prompts. we found that there is a lack of consistency across the  1) outputs of different models, and 2) outputs of a single model for simple variations of the prompt.
 \end{enumerate}

\section{Approach for Evaluation}
In this paper, we use the recently proposed \dset dataset \cite{dumpala2024sugarcrepe++} to evaluate the sensitivity of generative VLMs to lexical and semantic alterations. Each sample in the \dset dataset consists of an image and a triplet of captions (two positive captions, P$_1$ and P$_2$, and one negative caption, N). The two positive captions (P$_1$ and P$_2$) are lexically different but semantically similar, while the negative caption (N) is lexically closer to P$_1$ but semantically different from both P$_1$ and P$_2$. Additionally, \dset contains five different subsets, each created by replacing or swapping objects, attributes, and relations. Examples from the dataset are provided in Figure \ref{fig:dset} in the Appendix. For further details, refer to \citet{dumpala2024sugarcrepe++}. 

In this paper, we evaluate the ability of different generative VLMs (BLIP, BakLLaVA, and GPT-4o) to understand lexical and semantic alterations using the \dset dataset. We prompt these generative VLMs as follows: <Prompt> <Image><Caption1><Caption2><Caption3>, where <Prompt> refers to the query used to prompt the VLMs. Here, we use multiple variants of prompts that are semantically similar but lexically different. For instance, Table \ref{tab:bakllava_prompts} shows the variants of prompts used for BakLLaVA. The <Image>, <Caption1>, <Caption2>, and <Caption3> are samples from the \dset dataset. Each of <Caption1>, <Caption2>, and <Caption3> can be either P$_1$, P$_2$ or N. So, for each prompt, we report results on three variants obtained by reordering the positions of P$_1$, P$_2$ and N captions as follows: 1) N, P$_1$, P$_2$ 2) P$_1$, N, P$_2$ and 3) P$_1$, P$_2$, N.

\begin{table}[!tb]
\centering
\vspace{-0.37cm}
\caption{\small Different variants of the prompt used for prompt sensitivity analysis of BakLLaVA on \dset. '<image>' refers to the image corresponding to the captions provided as input to the BakLLaVA model.}
\label{tab:bakllava_prompts}
\vspace{0.2cm}
\resizebox{0.93\linewidth}{!}{
\begin{tabular}{cl}
\toprule
 & Prompt \\
\midrule
Prompt-1 & "USER: <image>\textbackslash n Which one of the three captions does not match the image? (1) <Caption1>; (2) <Caption2>; \\
& (3) <Caption3>; provide output as either (1), (2), or (3). Do not provide any explanation" +  "\textbackslash n ASSISTANT:" \\
\midrule
Prompt-2 & "USER: <image>\textbackslash n Which of the three captions does not correspond to the image? (1) <Caption1>; (2) <Caption2>; \\
& (3) <Caption3>; provide output as either (1), (2), or (3). Do not provide any explanation" +  "\textbackslash n ASSISTANT:" \\
\midrule
Prompt-3 & "USER: <image>\textbackslash n Which one of the three captions fails to correspond with the image? (1) <Caption1>; (2) <Caption2>; \\
& (3) <Caption3>; provide output as either (1), (2), or (3). Do not provide any explanation" +  "\textbackslash n ASSISTANT:" \\
\midrule
Prompt-4 & "USER: <image>\textbackslash n Which one of the three captions does not correspond to the image? (1) <Caption1>; (2) <Caption2>; \\
& (3) <Caption3>; provide output as either (1), (2), or (3). Do not provide any explanation" +  "\textbackslash n ASSISTANT:" \\
\midrule
Prompt-5 & "USER: <image>\textbackslash n Do any of these captions fail to correspond with the image? (1) <Caption1>; (2) <Caption2>; \\
& (3) <Caption3>; provide output as either (1), (2), or (3). Do not provide any explanation" +  "\textbackslash n ASSISTANT:" \\
\bottomrule
\end{tabular}}
\end{table}

\section{Results}
\subsection{Prompt Sensitivity of Generative VLMs}

\textbf{BakLLaVA:}
We use the five prompts listed in Table \ref{tab:bakllava_prompts} to evaluate BakLLaVA's sensitivity to prompt variations. These prompts are paraphrases of one another, conveying the same or similar semantic meaning. Table \ref{tab:bak_res} presents BakLLaVA's performance for each prompt across three variants (obtained by reorganizing the positions of the three options—P$_1$, P$_2$, and N). Two key observations can be made that are consistent across all the subsets of \dset: 1) We observe differences in performance when using paraphrases of the same prompt (lexically different but semantically identical). Moreover, no single prompt achieved the best performance across all \dset subsets. 2) Significant variations in BakLLaVA's performance on \dset were found for the same prompt, simply by reordering the positions of the three options. This highlights BakLLaVA's sensitivity to even minor changes in the input prompt.

\begin{table}[!tb]
\centering
\vspace{-0.37cm}
\caption{\small Prompt sensitivity analysis of BakLLaVA on \dset. We report the performance (Accuracy(\%)) for the three variants of each prompt (see Table \ref{tab:bakllava_prompts}) generated by reordering Positive 1 (P$_1$), Positive 2 (P$_2$) and Negative (N) captions. Overall best values are in bold, and prompt-level best values are underlined.}
\label{tab:bak_res}
\vspace{0.2cm}
\resizebox{0.93\linewidth}{!}{
\begin{tabular}{llrrrrr}
\toprule
Prompt & Variant & Swap Object & Swap Attribute & Replace Object & Replace Attribute & Replace Relation \\
\midrule
 & N, P$_1$, P$_2$ & 39.94 & 49.71 & 61.99 & 36.71 & 41.26 \\
Prompt-1 & P$_1$, N, P$_2$ & 33.52 & 47.18 & 59.41 & 41.50& 37.45 \\
 & P$_1$, P$_2$, N & \underline{68.08} & \underline{78.12}& \textbf{83.23}& \underline{73.84}& \underline{75.96}\\
 \midrule
 & N, P$_1$, P$_2$ & 31.72 & 30.73 & \underline{47.64}& 23.98 & 35.29\\
Prompt-2 & P$_1$, N, P$_2$ & 26.15 & 35.41 & 45.33& 24.52 & 32.09\\
 & P$_1$, P$_2$, N & \underline{64.12}& \underline{59.16}& 41.71 & \underline{53.42}& \underline{62.94}\\
 \midrule
 & N, P$_1$, P$_2$ & 36.94 & 40.33 & 46.91 & 37.31 & 48.61\\
Prompt-3 & P$_1$, N, P$_2$ & 30.61 & 36.05 & 51.85 & 44.23 & 54.97\\
 & P$_1$, P$_2$, N & \textbf{86.95}& \underline{67.72}& \underline{73.73}& \textbf{79.95}& \textbf{81.65}\\
 \midrule
 & N, P$_1$, P$_2$ & 47.53& 50.83 & 30.33 & 38.62 & 24.32 \\
Prompt-4 & P$_1$, N, P$_2$ & 51.26& 54.46 & 42.61 & 32.44 & 46.03 \\
 & P$_1$, P$_2$, N & \underline{75.11}& \textbf{81.83}& \underline{55.51}& \underline{64.34}& \underline{71.56}\\
 \midrule
  & N, P$_1$, P$_2$ &  \underline{54.19}&  \underline{61.54}&  \underline{53.14}&  \underline{60.71}&  \underline{67.42}\\
Prompt-5 & P$_1$, N, P$_2$ &  42.76&  49.78&  46.81&  48.11&  56.35\\
 & P$_1$, P$_2$, N &  38.06&  42.19&  47.68&  36.31&  49.30\\
 \bottomrule
\end{tabular}}
\end{table}

\textbf{GPT-4o: }
We evaluated the recently released GPT-4o~\footnote{\url{https://platform.openai.com/docs/models/gpt-4o}} ("o" for "omni") model on \dset using the prompts listed in Table \ref{tab:prompts_gpt4o}. Of the three prompts, Prompt-1 and Prompt-2 are paraphrases of each other and are structured as 4-class problems (where the output is (1), (2), (3), or "none"), while Prompt-3 is a 3-class problem, requiring the model to choose between (1), (2), or (3).

Table \ref{tab:gpt_4o} presents GPT-4o's performance on \dset for the three different prompts (with three variants for each prompt by reordering the options). Performance differences between Prompt-1 and Prompt-2, which are paraphrases of each other, highlight GPT-4o's sensitivity to prompt structure. When asked to choose from four options ((1), (2), (3), or "none"), GPT-4o had difficulty identifying the negative caption (a caption that does not correspond to the image). In contrast, it performed better at identifying the negative caption when asked to choose from three options ((1), (2), or (3)). Additionally, similar to BakLLaVA, GPT-4o's performance varied significantly for the same prompt simply by reordering the positions of the three options (P$_1$, P$_2$, and N). Moreover, these observations are consistent across all subsets of \dset. 
Refer to Section \ref{sec:blip} in Appendix for the results of the prompt sensitivity analysis on BLIP.

\begin{table}[!htb]
\vspace{-0.3cm}
\caption{\small Prompt sensitivity evaluation of GPT-4o on \dset. We report the performance (Accuracy(\%)) for the three variants of each prompt (see Table \ref{tab:prompts_gpt4o}) generated by reordering the Positive 1 (P$_1$), Positive 2 (P$_2$) and Negative (N) captions. Overall best values are in bold, and prompt-level best values are underlined.}
\label{tab:gpt_4o}
\vspace{0.3cm}
\centering
\footnotesize
\resizebox{0.95\linewidth}{!}{
\begin{tabular}{llrrrrr}
\toprule
Prompt & Variant & Swap Object & Swap Attribute & Replace Object & Replace Attribute & Replace Relation \\
\midrule
 & N, P$_1$, P$_2$ & 46.93& \underline{73.36}& 91.64& \underline{87.94}& 69.06\\
Prompt-1 & P$_1$, N, P$_2$ & 49.58& 69.22& 85.03& 83.62& \underline{70.41}\\
 & P$_1$, P$_2$, N & \underline{53.74}& 66.17& \underline{92.11}& 80.29& 66.56\\
 \midrule
 & N, P$_1$, P$_2$ & 48.25& \underline{75.04}& \underline{90.82}& 84.9& \underline{71.19}\\
Prompt-2 & P$_1$, N, P$_2$ & 45.36& 72.55& 86.71& 82.06& 64.51\\
 & P$_1$, P$_2$, N & \underline{51.43}& 69.25& 83.21& \underline{86.32}& 58.69\\
 \midrule
 & N, P$_1$, P$_2$ & 67.61& \textbf{85.82}& \textbf{96.25}& \textbf{93.27}& \textbf{84.13}\\
Prompt-3 & P$_1$, N, P$_2$ & 65.13& 83.29& 94.53& 88.75& 79.24\\
 & P$_1$, P$_2$, N & \textbf{70.67}& 79.39& 91.3& 90.61& 78.52\\
 \bottomrule
\end{tabular}}
\end{table}

\subsection{Evaluating Prompt Consistency of VLMs}
To evaluate the consistency of generative VLM's outputs in response to input prompts, we conducted two sets of experiments: 1) \textbf{Prompt-level consistency:} This evaluates a single model's consistency when provided with multiple paraphrased prompts. 2) \textbf{Inter-model consistency:} This assesses the consistency of outputs across different VLMs, given the same prompt. We analyzed BakLLaVA and GPT-4o using both approaches, applying majority voting on the outputs to determine the final decision.

\textbf{Prompt-level consistency:} We combine the outputs (using majority voting) of all three variants of each prompt, as well as the three variants across all prompts (All Prompts). Tables \ref{tab:bak_cons} and \ref{tab:gpt_4o_cons} present the performance of BakLLaVA and GPT-4o for each prompt (by combining the outputs of the three variants), and for All Prompts (by combining the outputs across all prompts), respectively.  We observe a lack of consistency in the model's outputs, even for the same prompt, simply by reordering the positions of the options. This inconsistency is highlighted by the fact that combining the outputs of the three variants for a given prompt often resulted in the lowest performance. For BakLLaVA, this was evident in the following cases: Prompts 4 and 5 for Swap Object; Prompts 1, 2, 4, and 5 for Swap Attribute; Prompts 2 and 5 for Replace Object; Prompt 1 for Replace Attribute; and Prompts 1, 2, 3, and 5 for Replace Relation. Similarly, for GPT-4o, we observed this pattern for Prompts 1 and 3 in Swap Object; Prompts 1, 2, and 3 in Swap Attribute; Prompt 2 for Replace Object; and Prompts 1 and 3 in Replace Relation.

\textbf{Inter-model Consistency:} 
We combine the outputs of BakLLaVA for Prompt-2 and Prompt-5 (Table \ref{tab:bakllava_prompts}) with the outputs of GPT-4o for Prompt-3 (Table \ref{tab:prompts_gpt4o}). As shown in Table \ref{tab:inter_model}, we observe that there is no strong agreement between the outputs of the models.

\begin{table}[!tb]
\centering
\vspace{-0.3cm}
\caption{\small Prompt-level consistency of BakLLaVA on \dset. Here we provide performance in terms of Accuracy (\%) by taking majority voting on the outputs of the three variants of each prompt, and that of all the prompts (All Prompts). (L): performance lower than three variants of prompt. See Table \ref{tab:bak_cons_comp} for complete results.}
\label{tab:bak_cons}
\vspace{0.2cm}
\centering
\footnotesize
\resizebox{0.93\linewidth}{!}{
\begin{tabular}{llllll}
\toprule
 & \multicolumn{1}{l}{Swap Object} & \multicolumn{1}{l}{Swap Attribute} & \multicolumn{1}{l}{Replace Object} & \multicolumn{1}{l}{Replace Attribute} & \multicolumn{1}{l}{Replace Relation} \\
\midrule
Prompt-1 & 41.27 & 45.93 (L)& 63.15& 35.32 (L)& 35.74 (L)\\
prompt-2 &  35.62 &  28.51 (L)&  29.36 (L)&  27.64&  20.83 (L)\\
Prompt-3 &  34.66 &  44.32&  \textbf{57.49}&  51.28&  47.05 (L)\\
prompt-4 &  26.39 (L)&  \textbf{48.22} (L)&  35.78&  41.59&  31.69\\
Prompt-5 &  31.24 (L)&  40.55 (L)&  41.83 (L)&  39.62&  43.09 (L)\\
\midrule
All Prompts &  \textbf{43.75}&  46.91&  46.07&  \textbf{52.96}&  \textbf{57.36}\\
\bottomrule
\end{tabular}}
\end{table}

\begin{table}[!htb]
% \vspace{-0.3cm}
\caption{\small Prompt-level consistency of GPT-4o on \dset. Here we provide performance in terms of Accuracy (\%) by taking majority voting on the outputs of the three variants of each prompt, and that of all the prompts (All Prompts). (L): performance lower than three variants of prompt. See Table \ref{tab:gpt-4o_cons_comp} for complete results.}
\label{tab:gpt_4o_cons}
\vspace{0.2cm}
\centering
\footnotesize
\resizebox{0.91\linewidth}{!}{
\begin{tabular}{llllll}
\toprule
 & \multicolumn{1}{c}{Swap Object} & \multicolumn{1}{c}{Swap Attribute} & \multicolumn{1}{c}{Replace Object} & \multicolumn{1}{c}{Replace Attribute} & \multicolumn{1}{c}{Replace Relation} \\
\midrule
 % Human & 100.00 & 96.67 & 100.00 & 100.00 & 100.00 \\
 % \midrule
 Prompt-1 & 43.16 (L)& 65.69 (L) & 90.25& 86.32 & 65.57 (L)\\
 Prompt-2 & 46.64 & 67.34 (L)& 85.49& 81.07 (L)& 62.35\\
 Prompt-3 & 62.36 (L) & 78.54 (L) & 92.41 & 90.31 & 77.72 (L) \\ 
 \midrule
 All Prompts & 57.35& 72.81& 88.15& 85.81& 73.62\\ 
\bottomrule
\end{tabular}}
\end{table}

\begin{table}[!htb]
\centering
\vspace{-0.3cm}
\caption{\small Inter-model consistency of BakLLaVA and GPT-4o (Accuracy (\%)). See Table \ref{tab:inter_model_full} for complete results.}
\label{tab:inter_model}
\vspace{0.2cm}
\centering
\footnotesize
\resizebox{0.93\linewidth}{!}{
\begin{tabular}{llllll}
\toprule
 & \multicolumn{1}{l}{Swap Object} & \multicolumn{1}{l}{Swap Attribute} & \multicolumn{1}{l}{Replace Object} & \multicolumn{1}{l}{Replace Attribute} & \multicolumn{1}{l}{Replace Relation} \\
\midrule
 & 51.82 & 52.18 & 43.29 & 45.91 & 46.37\\
\bottomrule
\end{tabular}}
\end{table}

\section{Conclusions}
In this work, we evaluated the prompt sensitivity of generative VLMs (BakLLaVA, GPT-4o, and BLIP) using the \dset dataset. Our findings reveal significant inconsistencies in the models’ outputs both with multiple paraphrases of the same prompt and by simply reordering options within the same prompt. Furthermore, we find that naive majority voting over different orders of options does not consistently improve performance. Generative VLMs demonstrate inconsistent behavior both across paraphrased prompts and between different models. These results highlight the need for improved robustness against lexical variations in generative VLMs.

% \newpage
\bibliographystyle{abbrvnat}
\bibliography{references,refs-eem}

\newpage
\appendix

\section{Dataset Details}
\begin{figure}[!tb]
\hspace{0.82cm}
% \vspace{-0.3cm}
    \centering
    \includegraphics[width=0.92\linewidth]{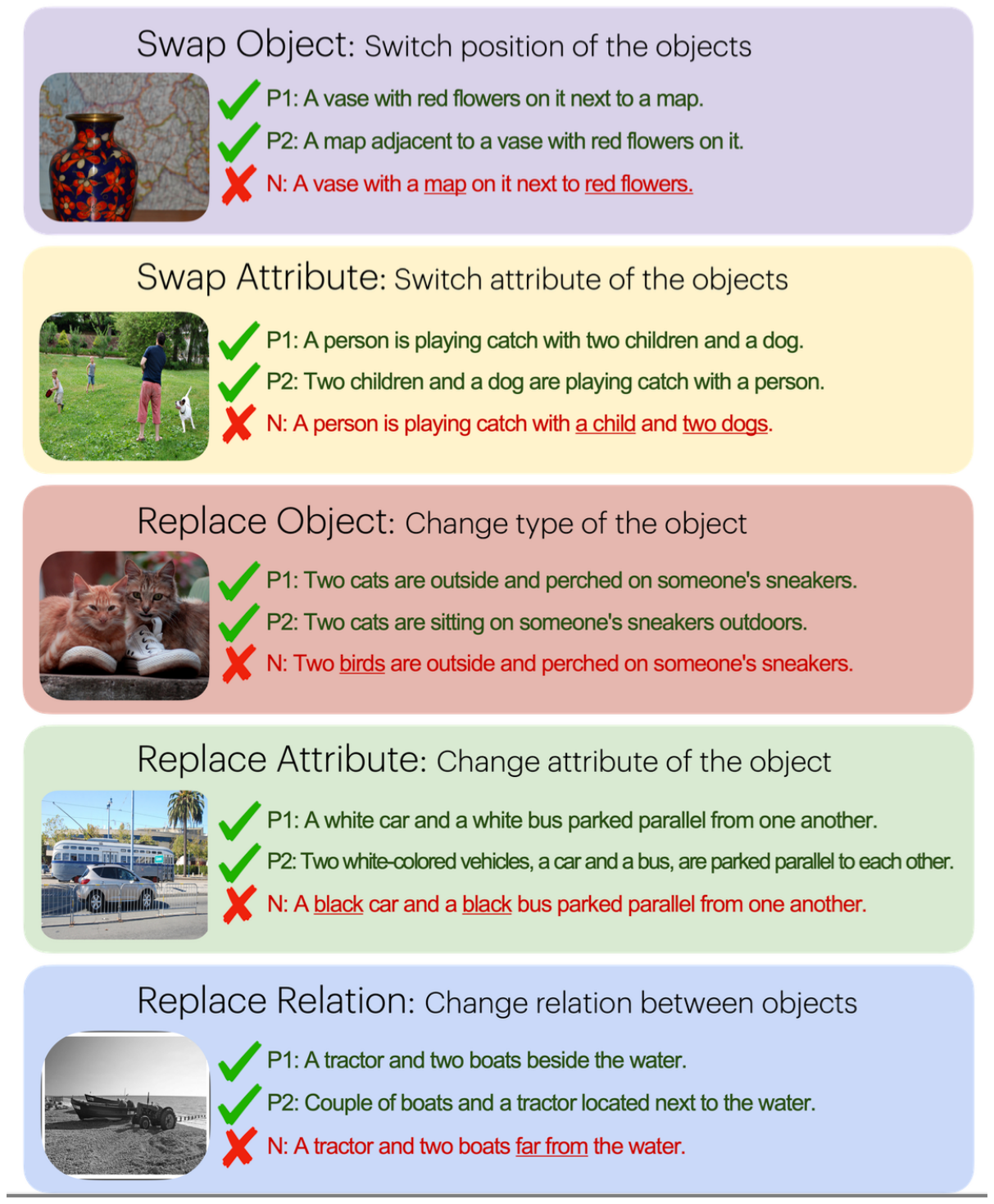}
    % \vspace{0.1cm}
    \caption{\small 
    %Figure shows an 
    Examples from \dset (SC++) dataset. $P_1$ and $P_2$ are semantically equivalent but lexically different while $N$ is semantically different than both $P_1$ and $P_2$ despite its lexical similarity with $P_1$.} 
    % The lexically altered second caption preserves the general semantics of the scene in the image, while the third caption contradicts the scene. These captions are referred to as hard positive and hard negative, respectively.
    \label{fig:dset}
    \vspace{-8pt}
\end{figure}

Figure \ref{fig:dset} provides examples of each subset from the \dset dataset. The distribution of samples in \dset dataset are provided in Table \ref{tab:sc_stats}.

\begin{table}[!b]
% \vspace{0.2cm}
\caption{\dset consists of 4757 examples with the following distribution of sample sizes 
% Examples for each negative type are available in Figure \ref{fig:example}.
}
\label{tab:sc_stats}
\vspace{0.2cm}
    \centering
    \begin{tabular}{cccccc}
    \toprule
    Swap Object & Swap Attribute& Replace Object& Replace Attribute& Replace relation & \textbf{Total}\\
    \midrule
    245& 666 & 1652 & 788 & 1406 &  4757  \\
    \bottomrule
    \end{tabular}
    % \vspace{-5pt}
\end{table}

\begin{table}[!htb]
\caption{\small Different variants of the prompt used for prompt sensitivity analysis of GPT-4o on \dset. '<image>' refers to the image corresponding to the captions provided as input to the BakLLaVA model.}
\label{tab:prompts_gpt4o}
\vspace{0.2cm}
\resizebox{1\linewidth}{!}{
\begin{tabular}{cl}
\toprule
Prompt-1 & "Do any of the following captions not match the image? (1) " + <Caption 1> + "; (2) " + <Caption 2> + "; \\
& (3) " + <Caption 3> +  "; provide output as (1), (2), (3) or none" \\
\midrule
Prompt-2 & "Do any of these captions fail to correspond with the image? (1) " + <Caption 1> + "; (2) " + <Caption 2> + "; \\
& (3) " + <Caption 3> +  "; provide output as (1), (2), (3) or none" \\
\midrule
Prompt-3 & "Do any of these captions fail to correspond with the image? (1) " + <Caption 1> + "; (2) " + <Caption 2> + "; \\
& (3) " + <Caption 3> +  "; provide output as (1), (2), (3) or none" \\
\bottomrule
\end{tabular}}
\end{table}

% \begin{itemize}
% \item \textbf{Prompt-2:} "Do any of these captions fail to correspond with the image? (1) " + <Negative Caption> + "; (2) " + <Positive Caption 1> + "; (3) " + <Positive caption 2> +  "; provide output as (1), (2), (3) or none"
% \item \textbf{Prompt-3:} "Do any of these captions fail to correspond with the image? ((1) " + <Negative Caption> + "; (2) " + <Positive Caption 1> + "; (3) " + <Positive caption 2> +  "; provide output as (1), (2) or (3)"
% \end{itemize}

\section{Additional Results}

\textbf{Evaluation:} We provide both image and a prompt to the BakLLaVA and GPT-4o, and receive the output from the models. Based on the response, we compute the performance i.e., it is a hit if the model outputs the correct position of the negative caption else a miss. Performance is reported in terms of Accuracy (\%), where accuracy is computed as the ratio of the \#hits/(\#hits + \#misses).

\subsection{BLIP}
\label{sec:blip}
We evaluate the BLIP model using the prompts listed in Table~\ref{blip_prompts}. As shown in the table, we experiment with multiple paraphrased versions of two different prompts (Prompt-1 and Prompt-2), which are slightly different from each other. Significant differences in BLIP's performance can be observed, even for paraphrases of the same prompt—this is evident for both Prompt-1 and Prompt-2. Additionally, no single prompt achieved the best performance across all subsets of \dset.

\begin{table}[!htb]
\centering
\caption{\small Prompt-sensitivity analysis of BLIP (a generative VLM) using \dset. Results show that simple paraphrases of the input prompt significantly affect model performance. [Image]: Corresponding image provided to the image encoder along with the prompt. We provide three separate inputs for each example by replacing '<Caption>' with either Positive 1, Positive 2, or Negative captions.}
\label{blip_prompts}
\vspace{0.2cm}
\resizebox{1\linewidth}{!}{
\begin{tabular}{clrrrrr}
\toprule
 & \multicolumn{1}{c}{Input Prompts} & \multicolumn{1}{l}{S. Obj} & \multicolumn{1}{l}{S. Att} & \multicolumn{1}{l}{R. Obj} & \multicolumn{1}{l}{R. Att} & \multicolumn{1}{l}{R. Rel} \\
 \midrule
 & \begin{tabular}[c]{@{}l@{}} [Image] Does the following caption match the image: \\ \textless{}Caption\textgreater{}? Provide 'yes' or 'no' as response\end{tabular} & 6.1 & 18.12 & 48.09 & 34.64 & 16.06 \\
\cline{2-7}
 \begin{tabular}[c]{@{}l@{}}Prompt-1 \\ Paraphrases\end{tabular} & \begin{tabular}[c]{@{}l@{}}Does the caption match the image:\\  \textless{}Caption\textgreater{}? Provide 'yes' or 'no' as response\end{tabular} & 7.34 & 20.87 & 45.46 & 37.56 & 18.55 \\
 \cline{2-7}
 & \begin{tabular}[c]{@{}l@{}} [Image] Do the image and the following caption match: \\ \textless{}Caption\textgreater{}? Provide 'yes' or 'no' as response\end{tabular} & 8.21 & 12.46 & 40.86 & 29.06 & 11.95 \\
 \midrule
 \begin{tabular}[c]{@{}l@{}} Prompt-2 \\ Paraphrases\end{tabular}& \begin{tabular}[c]{@{}l@{}}Does the caption correctly describe the image:\\  \textless{}Caption\textgreater{}? Provide 'yes' or 'no' as response\end{tabular} & 6.53 & 24.47 & 51.51 & 41.16 & 23.25 \\
\cline{2-7}
 & \begin{tabular}[c]{@{}l@{}}[Image] Is the caption an accurate description of the image: \\ \textless{}Caption\textgreater{}? Provide 'yes' or 'no' as response.\end{tabular} & 7.34 & 23.27 & 53.39 & 42.89 & 25.1 \\ 
 \bottomrule
\end{tabular}}
\end{table}

\begin{table}[!htb]
\centering
\vspace{-0.37cm}
\caption{\small Prompt-level consistency of BakLLaVA on \dset. Here we provide performance in terms of Accuracy (\%) by taking majority voting on the outputs of the three variants of each prompt, and that of all the prompts (All Prompts). MV refers to majority voting over all variants of a prompt. (L): performance of MV lower than three variants of the prompt.}
\label{tab:bak_cons_comp}
\vspace{0.2cm}
\vspace{0.2cm}
\resizebox{1\linewidth}{!}{
\begin{tabular}{lllllll}
\toprule
Prompt & Variant & Swap Object & Swap Attribute & Replace Object & Replace Attribute & Replace Relation \\
\midrule
 & N, P$_1$, P$_2$ & 39.94 & 49.71 & 61.99 & 36.71 & 41.26 \\
Prompt-1 & P$_1$, N, P$_2$ & 33.52 & 47.18 & 59.41 & 41.50& 37.45 \\
 & P$_1$, P$_2$, N & \underline{68.08} & \underline{78.12}& \textbf{83.23}& \underline{73.84}& \underline{75.96}\\
 & MV & 41.27& 45.93 (L)& 63.15& 35.32 (L)& 35.74 (L)\\
 \midrule
 & N, P$_1$, P$_2$ & 31.72 & 30.73 & \underline{47.64}& 23.98 & 35.29\\
Prompt-2 & P$_1$, N, P$_2$ & 26.15 & 35.41 & 45.33& 24.52 & 32.09\\
 & P$_1$, P$_2$, N & \underline{64.12}& \underline{59.16}& 41.71 & \underline{53.42}& \underline{62.94}\\
 & MV & 35.62&  28.51 (L)&  29.36 (L)&  27.64&  20.83 (L)\\
 \midrule
 & N, P$_1$, P$_2$ & 36.94 & 40.33 & 46.91 & 37.31 & 48.61\\
Prompt-3 & P$_1$, N, P$_2$ & 30.61 & 36.05 & 51.85 & 44.23 & 54.97\\
 & P$_1$, P$_2$, N & \textbf{86.95}& \underline{67.72}& \underline{73.73}& \textbf{79.95}& \textbf{81.65}\\
 & MV & 34.66&  44.32&  57.49 &  51.28&  47.05 (L)\\
 \midrule
 & N, P$_1$, P$_2$ & 47.53& 50.83 & 30.33 & 38.62 & 24.32 \\
Prompt-4 & P$_1$, N, P$_2$ & 51.26& 54.46 & 42.61 & 32.44 & 46.03 \\
 & P$_1$, P$_2$, N & \underline{75.11}& \textbf{81.83}& \underline{55.51}& \underline{64.34}& \underline{71.56}\\
 & MV & 26.39 (L)&  48.22 (L) &  35.78&  41.59&  31.69\\
 \midrule
  & N, P$_1$, P$_2$ &  \underline{54.19}&  \underline{61.54}&  \underline{53.14}&  \underline{60.71}&  \underline{67.42}\\
Prompt-5 & P$_1$, N, P$_2$ &  42.76&  49.78&  46.81&  48.11&  56.35\\
 & P$_1$, P$_2$, N &  38.06&  42.19&  47.68&  36.31&  49.30\\
 & MV & 31.24 (L)&  40.55 (L)&  41.83 (L)&  39.62&  43.09 (L)\\
 \midrule
All Prompts & MV & 43.75 &  46.91&  46.07&  52.96& 57.36\\
 \bottomrule
\end{tabular}}
\end{table}

\begin{table}[!htb]
\vspace{-0.37cm}
\caption{\small Prompt-level consistency of BakLLaVA on \dset. Here we provide performance in terms of Accuracy (\%) by taking majority voting on the outputs of the three variants of each prompt, and that of all the prompts (All Prompts). MV refers to majority voting over all variants of a prompt. (L): performance of MV lower than three variants of the prompt.}
\label{tab:gpt-4o_cons_comp}
\vspace{0.3cm}
\centering
\footnotesize
\resizebox{0.95\linewidth}{!}{
\begin{tabular}{lllllll}
\toprule
Prompt & Variant & Swap Object & Swap Attribute & Replace Object & Replace Attribute & Replace Relation \\
\midrule
 & N, P$_1$, P$_2$ & 46.93& \underline{73.36}& 91.64& \underline{87.94}& 69.06\\
Prompt-1 & P$_1$, N, P$_2$ & 49.58& 69.22& 85.03& 83.62& \underline{70.41}\\
 & P$_1$, P$_2$, N & \underline{53.74}& 66.17& \underline{92.11}& 80.29& 66.56\\
 & MV & 43.16 (L)& 65.69 (L) & 90.25& 86.32 & 65.57 (L)\\
 \midrule
 & N, P$_1$, P$_2$ & 48.25& \underline{75.04}& \underline{90.82}& 84.90& \underline{71.19}\\
Prompt-2 & P$_1$, N, P$_2$ & 45.36& 72.55& 86.71& 82.06& 64.51\\
 & P$_1$, P$_2$, N & \underline{51.43}& 69.25& 83.21& \underline{86.32}& 58.69\\
 & MV & 46.64 & 67.34 (L)& 85.49& 81.07 (L)& 62.35\\
 \midrule
 & N, P$_1$, P$_2$ & 67.61& \textbf{85.82}& \textbf{96.25}& \textbf{93.27}& \textbf{84.13}\\
Prompt-3 & P$_1$, N, P$_2$ & 65.13& 83.29& 94.53& 88.75& 79.24\\
 & P$_1$, P$_2$, N & \textbf{70.67}& 79.39& 91.30& 90.61& 78.52\\
 & MV & 62.37 (L) & 78.54 (L) & 92.41 & 90.31 & 77.72 (L) \\
 \midrule
 All Prompts & MV & 57.35& 72.81& 88.15& 85.81& 73.62\\ 
 \bottomrule
\end{tabular}}
\end{table}

\begin{table}[!htb]
\centering
\vspace{-0.37cm}
\caption{\small Inter-model consistency of BakLLaVA and GPT-4o (Accuracy (\%)).}
\label{tab:inter_model_full}
\vspace{0.2cm}
\centering
\footnotesize
\resizebox{0.93\linewidth}{!}{
\begin{tabular}{llllllll}
\toprule
 Model & Prompt& Variant & \multicolumn{1}{l}{Swap Object} & \multicolumn{1}{l}{Swap Attribute} & \multicolumn{1}{l}{Replace Object} & \multicolumn{1}{l}{Replace Attribute} & \multicolumn{1}{l}{Replace Relation} \\
\midrule
& & N, P$_1$, P$_2$ & 31.72 & 30.73 & \underline{47.64}& 23.98 & 35.29\\
& Prompt-2 & P$_1$, N, P$_2$ & 26.15 & 35.41 & 45.33& 24.52 & 32.09\\
BakLLaVA & & P$_1$, P$_2$, N & \underline{64.12}& \underline{59.16}& 41.71 & \underline{53.42}& \underline{62.94}\\
 \cmidrule{2-8}
  & & N, P$_1$, P$_2$ &  \underline{54.19}&  \underline{61.54}&  \underline{53.14}&  \underline{60.71}&  \underline{67.42}\\
& Prompt-5 & P$_1$, N, P$_2$ &  42.76&  49.78&  46.81&  48.11&  56.35\\
& & P$_1$, P$_2$, N &  38.06&  42.19&  47.68&  36.31&  49.30\\
\midrule
& & N, P$_1$, P$_2$ & 67.61& \textbf{85.82}& \textbf{96.25}& \textbf{93.27}& \textbf{84.13}\\
GPT-4o & Prompt-3 & P$_1$, N, P$_2$ & 65.13& 83.29& 94.53& 88.75& 79.24\\
& & P$_1$, P$_2$, N & \textbf{70.67}& 79.39& 91.30& 90.61& 78.52\\
\midrule
 & & N, P$_1$, P$_2$ & 46.53 & 54.43 & 41.95 & 44.76 & 49.35 \\
Inter-Model & & P$_1$, N, P$_2$ & 35.21 & 40.27 & 39.07 & 46.85 & 47.51 \\
& & P$_1$, P$_2$, N & 49.67 & 49.64 & 45.34 & 49.23 & 54.38  \\
\cmidrule{2-8}
& & All & 51.82 & 52.18 & 43.29 & 45.91 & 46.37 \\
\bottomrule
\end{tabular}}
\end{table}

\end{document}